\documentclass{article}
\usepackage{spconf,amsmath,graphicx,hyperref}

\usepackage{amsmath,amssymb,amsfonts, enumitem, fancyhdr, color, comment, graphicx, environ,changepage, mathrsfs}
\usepackage{quiver,tikz-cd}
\usepackage{hyperref}

\def\bx{\mathbf{x}}

\def\base{\tilde{\mathbf{x}}}

\def\bgamma{\boldsymbol{\gamma}}

\title{Higher-Order Feature Attribution: Bridging Statistics, Explainable AI, and Topological Signal Processing}
\name{
Kurt Butler$^{1,2}$, Guanchao Feng$^3$, and Petar M. Djuri{\'c}$^3$
\thanks{This work was supported by NSF under Award 2212506, the UKRI AI programme, and the Engineering and Physical Sciences Research Council, for CHAI - Causality in Healthcare AI Hub [grant number EP/Y028856/1]. Accepted to the IEEE International Conference on Acoustics, Speech and Signal Processing (ICASSP) 2026. © 2026 IEEE.}}
\address{
$^1$School of Engineering, The University of Edinburgh, Edinburgh, UK
\\
$^2$Causality in Healthcare AI Hub (CHAI), UK
\\
$^3$Department of Electrical and Computer Engineering, Stony Brook University, Stony Brook, NY, USA
}

\begin{document}
%\ninept
%
\maketitle
\begin{abstract}
Feature attributions are post-training analysis methods that assess how various input features of a machine learning model contribute to an output prediction. Their interpretation is straightforward when features act independently, but it becomes less clear when the predictive model involves interactions, such as multiplicative relationships or joint feature contributions.
In this work, we propose a general theory of higher-order feature attribution, which we develop on the foundation of Integrated Gradients (IG).  This work extends existing frameworks in the literature on explainable AI. When using IG as the method of feature attribution, we discover natural connections to statistics and topological signal processing. We provide several theoretical results that establish the theory, and we validate our theory on a few examples. 
\end{abstract}
\section{Introduction}
Explainable artificial intelligence (XAI) is a discipline that seeks to develop tools that can extract insights from \textit{black-box} predictive models regarding how they make predictions. Such explanations are of great interest when these models are used for scientific purposes \cite{gunning2019xai} or high-stakes decision making \cite{rudin2019stop}. While methods such as deep neural networks, transformers, and kernel machines exploit techniques that are interpretable in their own right, the connections between these models are not always apparent. Techniques used to explain the behavior of one model, such as neuron activations or kernel weights, may not make sense for other model architectures.  
XAI attempts to find techniques for explaining models that can be applied to any model in a unified way, without requiring specific assumptions about the architecture of the model under study.

Feature attributions solve the XAI problem by providing a way to \textit{attribute} the prediction to the input features, meaning that  one quantifies the contribution of each feature to the change in the output prediction \cite{sundararajan2020many}. For example, in \cite{heskes2020causal}, an XGBoost model is used to predict the number of bike rentals using covariates such as wind speed and temperature. The authors compute the attributions of this model, which quantify how much the particular wind speed or temperature of a given day has raised or lowered the total number of predicted bike rentals.
Even in this simple example, there lies an issue. The contributions of features such as wind speed or temperature may not be additively separable. For example, if the model changes its sensitivity to wind speed depending on the temperature, then the contributions of each feature are mutually intertwined. While feature attributions can still be computed in this case, this \textit{interaction} between features cannot be detected from the attributions alone.

\begin{figure}
    \centering
    \includegraphics[width=0.99\linewidth]{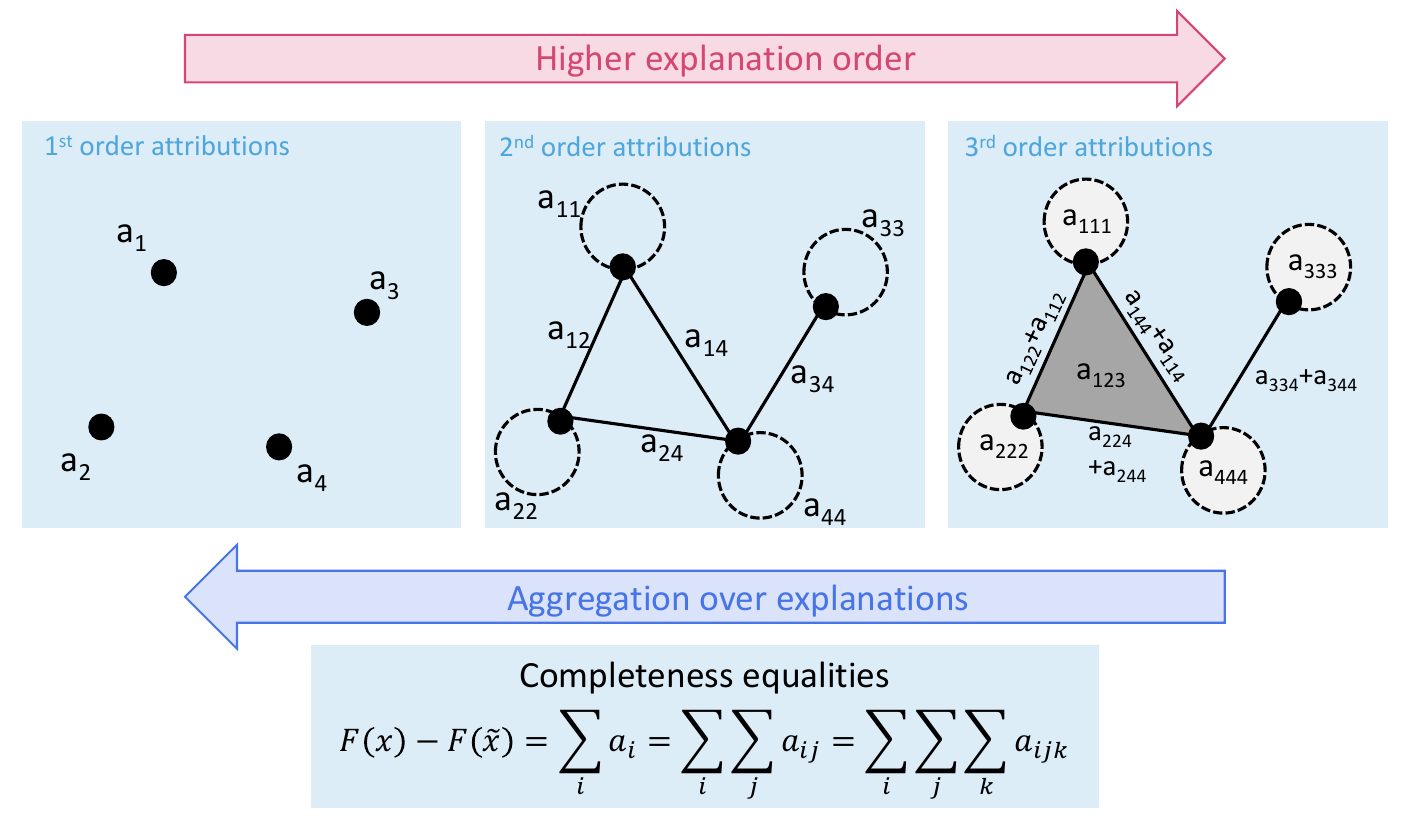}
    \caption{A visualization of our proposed approach. In standard feature attribution, a prediction is decomposed into attributions $a_i$, which quantify the contribution of each input feature in the predictive model. Repeating this procedure yields increasingly refined information, which can be arranged geometrically as graphs or their higher order analogues.}
    \label{fig:overview}
\end{figure}

To clarify the connection between feature attribution and interaction, we propose a general theory of higher-order feature attribution. In this work, we use Integrated Gradients (IG) as our feature attribution method of choice \cite{sundararajan2017axiomatic}, and we show that our theory generalizes the framework of Integrated Hessians \cite{janizek2021explaining}. Our theory also admits a graphical representation of feature attributions (Fig. \ref{fig:overview}), where graph signals can be used as visual explanations of a prediction. Overall, our framework eases the interpretation and manipulation of feature attributions as mathematical objects and provides new insights into how to interpret them.

The remainder of the paper is organized as follows. In Sec. \ref{sec:background}, we briefly introduce necessary concepts from statistics and machine learning to contextualize this work. In Sec. \ref{sec:theory}, we define our notion of higher-order feature attributions. We discuss several experiments in Sec. \ref{sec:experiments}. We discuss related work and then conclude the article in Sec. \ref{sec:conclusion}

\section{Background}
\label{sec:background}
In this section, we provide a brief primer on the concepts of interaction and feature attribution.

\textbf{Interactions} quantify the extent to which the strength of influence from an input variable to the output prediction depends on another input variable \cite{vanderweele2014tutorial}, specifically in the context of regression. Importantly, this notion of interaction does not refer to the correlation between the input features themselves, but rather to their behavior within a predictive model. The prototypical example is a monomial function, such as $f(x_1,x_2) = 3x_1 x_2$. In this example, sensitivity of the function's output to the variable $x_1$ depends on the value of $x_2$, and vice versa. As such, one cannot compute a measure of the influence of $x_1$ on $f(x_1,x_2)$ without the stratification or marginalization of $x_2$ \cite{butler2024measuring}. 
Practitioners of statistics are often interested in modeling or detecting interactions within their own domains of study, such as pharmacology \cite{huang2007drug}. Techniques used to quantify and detect the presence of interactions include analysis-of-variance (ANOVA) methods \cite{leys2010nonparametric} and Sobol indices \cite{sobol2001global}, although these methods are not immediately related to feature attributions.

\textbf{Feature attributions} are a measure of the contribution of each input feature in a predictive model to the output. Consider a regression model of the form
\begin{equation}
    y = f(\bx) + \epsilon.
    \label{eq:regression_model}
\end{equation}
The input vector $\bx \in \mathbb{R}^D$ represents input data, which might consist of tabular data, images, etc. The predictive model is represented as a mathematical function $f$ that takes in $\bx$ and outputs an estimate of the target variable $y \in \mathbb{R}$. The residual $\epsilon$ represents the prediction error and is not relevant here. 

In applied settings, a natural question is ``How does each covariate $x_i$ in $\bx$ contribute to my prediction $f(\bx)$?''. For example, suppose that $\bx$ represents a hospital patient's clinical profile, and $y$ represents the probability that they will develop Alzheimer's disease. Furthermore, suppose that for this specific patient, the predicted risk $f(\bx)$ is higher than a standard patient's risk. It is natural to ask which features in the patient's profile have contributed to this specific prediction. 

For linear models, feature attributions have a canonical definition. If 
$f(\bx) = \beta_1 x_1 + \beta_2 x_2 + \cdots + \beta_D x_D,$
then the contribution of the $i$-th feature to the relative change in prediction $f(\bx)-f(\base)$ is given by 
\begin{equation}
    \label{eq:linear_feature_attr}
    a_i = \beta_i (x_i - \tilde{x}_i),
\end{equation}
which we call the attribution to feature $x_i$. By construction, feature attributions are designed to satisfy the completeness property,
\begin{equation}
    \label{eq:completeness}
    f(\bx) - f(\base) = \sum_i a_i.
\end{equation}
The result of this is that $a_i$ encodes the contribution of each input feature to the relative change in a prediction $f(\bx)$ from a given baseline prediction $f(\base)$.

We note that this definition of $a_i$ assumes that $f$ is a linear function of $\bx$. To extend feature attributions beyond linear models to general nonlinear functions, a new definition is required. There are several possible approaches to this, such as Shapley values and LIME \cite{fumagalli2023shap,ribeiro2016should}. 
\textbf{Integrated Gradients} (IG) is a particular method of feature attribution that satisfies key axioms of attribution methods \cite{sundararajan2020many}
\begin{equation}
    \label{eq:IntegratedGradients}
    a_i  = (x_i - \tilde{x}_i) \int_0^1 \frac{\partial f(\bgamma(\bx,t))}{\partial x_i} dt,
\end{equation}
where $\bgamma(\bx,t) = t \bx + (1-t) \base$. 

\textbf{Linear operator perspective}.
In the standard XAI literature, $a_i$ is considered a constant number given the input location $\bx$. However, we may also view $a_i$ as a function of $\bx$. In this view, the process of taking an attribution is understood as starting with a predictive function $f$, and then applying an attribution operator $A_i$ to yield a new function $a_i(\bx) = A_i f(\bx)$ \cite{butler2024explainable}. From \eqref{eq:IntegratedGradients}, it can be seen that the operator $A_i$ is a linear operator on a space of functions. As a result, attribution operators can be composed naturally, similar to derivative operators, which leads to our notion of higher-order attributions. The analysis in this paper is limited to IG attribution operators, but operator theories for other feature attribution definitions are topics left for future work.

\section{Second-Order Attribution Theory}
\label{sec:theory}
We define second-order attributions in analogy to second-order derivatives. That is, we define a second-order attribution by the composition of two attribution operators:
\begin{equation}
    a_{ij}(\bx) = A_i A_j f(\bx).
\end{equation}
Intuitively, if feature attribution is a process that decomposes a prediction into the contributions of each feature, then the above composition is a second decomposition that corresponds to the contributions of \textit{pairs of features}.

When using the IG definition for feature attributions \eqref{eq:IntegratedGradients}, one can derive analytical expressions for $a_{ij}$. It can be seen that the following expressions coincide with those of the Integrated Hessians framework introduced in \cite{janizek2021explaining}.
If $i \neq j$, then the \textit{ mixed attributions} ($i \neq j$) are given by
\begin{equation}
    \label{eq:mixedattr}
    a_{ij} = A_i A_j f(\bx) = 
     \Delta x_i \Delta x_j \int_0^1 \int_0^1  \frac{\partial^2 f(\bgamma(\bx,st))}{\partial x_j \partial x_i}  s t\,  dsdt,
\end{equation}
and the \textit{repeated attributions} ($i=j$) are given by
\begin{align}
    \label{eq:squaredattr}
    a_{ii} = A_i A_i f(\bx)  &= \Delta x_i  \int_0^1 \int_0^1 \frac{\partial f(\bgamma(\bx,st))}{\partial x_i} dsdt
    \notag \\ &  +
     \Delta x_i^2 \int_0^1 \int_0^1 \frac{\partial^2 f(\bgamma(\bx,st))}{\partial x_i^2} st \, ds dt.
\end{align}
The appearance of the term $\bgamma(\bx,st)$ is given by the fact that $\bgamma(\bgamma(\bx,s),t)=\bgamma(\bx,st)$, which can be verified via algebra.

We now note several useful properties of the second-order attributions.
Firstly, since each $A_i$ is linear, the second-order attributions $A_i A_j$ are automatically linear as well. 
Also, attribution operators satisfy a \textbf{symmetry} property: $A_i A_j f(\bx) = A_j A_i f(\bx)$. Like derivatives, the order of application does not matter.

 \textbf{Marginalization:}
We can recover the first order attributions from their second-order counterparts according to
\begin{equation}
    A_i f(\bx) = \sum_{j=1}^D A_i A_j f(\bx).
\end{equation}
 \textbf{Completeness:}
Following from marginalization, we yield another version of the completeness property \eqref{eq:completeness}:
\begin{equation}
    f(\bx) - f(\base) =\sum_{i,j} A_i A_j f(\bx) = \sum_{i,j} a_{ij}.
\end{equation}
\textbf{Additive models.}
If $f$ is an additive model, i.e., $f(\bx) = f_1(x_1) + \cdots + f_D(x_D)$, then 
\begin{align}
    A_i^2 f(\bx) &= A_if(\bx) =  f_i(x_i) - f_i(\tilde{x}_i), &  \\
    A_i A_j f(\bx) &\equiv 0, \qquad \qquad \text{if }i \neq j.
    \label{eq:mixed_attr_zero_lin_model}
\end{align}

% \subsection{Higher Orders}
The extension of the theory to higher orders is again clear from the analogy to derivatives. Third-order attributions are computed through the application of three attribution operators. Hence, $a_{ijk} = A_i A_j A_k f(\bx)$. As stated above, the marginalization property allows us to relate explanations of different orders via summation over various indices. 

\subsection{Connections to topological signal processing}
The operator theory of higher-order attributions is also naturally related to concepts from graph and topological signal processing \cite{Barbarossa2020topological, mateos2019connecting}. In short, for any prediction of the model $f(\bx)$, we can represent a second-order (or higher-order) explanation as a signal over a graph (or a simplicial complex) that has particular algebraic properties. 

\textbf{Tensor representation}. Our choice to denote first, second, and third attributions as $a_{i}, a_{ij},$ and $a_{ijk}$ is suggestive. One can record attributions of order $L$ into a tensor of order $L$. The marginalization property  suggests that these tensors of different orders are related by tensor contraction. Storing attributions in a tensor is natural if one considers representing these explanations on a computer. 

\textbf{Topological representation}.
As a starting point, we consider second-order attributions. Consider a graph $\mathcal{G}$ whose nodes correspond to features $x_i$, and there exists an edge $x_i \to x_j$ if there is an interaction effect between $x_i$ and $x_j$ (that is, if $a_{ij} \neq 0$). Self-loops are included by default in $\mathcal{G}$. Explanations of predictions correspond to signals defined over this graph:
First-order attributions $a_i$ correspond to a graph signal over the nodes $\mathcal{G}$.
Second-order attributions $a_{ij}$ correspond to a graph signal over the edges in $\mathcal{G}$.
 The completeness property implies a relation between the edge signals and node signals. Namely, each node signal equals the sum of the adjacent edge signals.

Moving from second to higher order attributions, we also need to consider higher-dimensional extensions of graphs. For these cases, there is some flexibility in selecting an approach. 
One approach is to consider simplicial complexes \cite{Barbarossa2020topological}, which can be facilitated by using multiplicity encodings or by aggregating multiple attributions into common edges; e.g., the edge $(i,j)$ represents $a_{ijj} + a_{iij}$ in the third order case. Aggregation in this case preserves the property that summation over adjacent elements reproduces lower order explanations, and it is not necessarily intuitive to imagine the difference between $a_{112}$ and $a_{122}$ conceptually, so aggregation may be useful for visualizing third order effects. Exploring the advantages or disadvantages of each representation is left as a topic for future work.

\section{Experiments}
\label{sec:experiments}
We now consider some empirical evidence for our approach. 

\subsection{Synthetic data experiment}
\begin{figure}[h]
    \centering
    \includegraphics[width=0.8\linewidth]{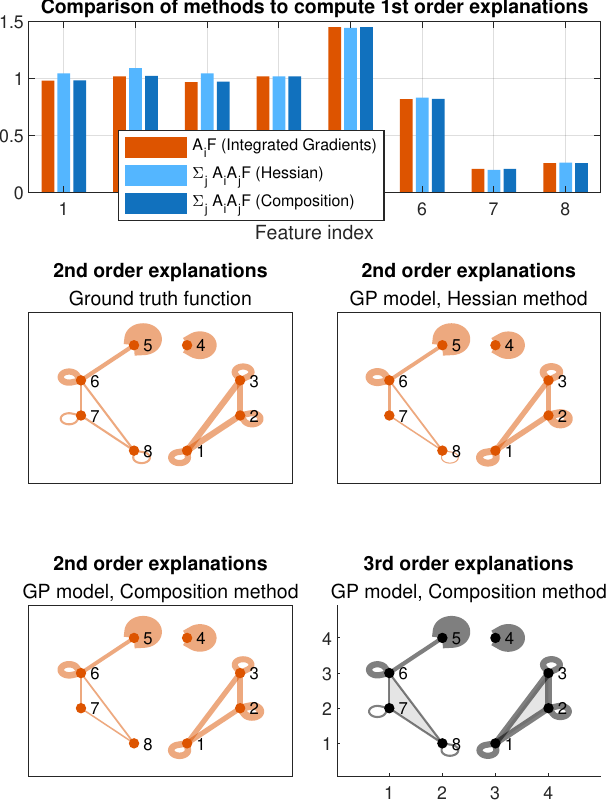}
    \caption{An unknown function is estimated from data using a GPR model. We visualize separately first, second and third order explanations using various approaches. We also show a `ground truth' explanation, for a perfectly estimated function.}
    \label{fig:synthetic}
\end{figure}

In Fig. \ref{fig:synthetic}, we analyze data generated by the following generative model. 
\begin{align}
    x_i &\sim \mathcal{U}(0,1)  \qquad \qquad \qquad i=1,...,8\\
    y &= f(\bx) + 0.1 \mathcal{N}(0,1)
    \\ f(\bx)
    &= 3x_1 x_2 x_3 + x_4  + x_5 + x_5 x_6 + x_6 x_7 x_8
    \label{eq:synthpoly}
\end{align}
Because there is a ground truth function, there is a notion of ground truth interactions that are present in the system. Thus, the goal of this experiment is to verify that we can recover this interaction structure.

We observe 500 samples from this generative process and train a Gaussian process regression (GPR) model to fit the data. 
In Fig. \ref{fig:synthetic}, we compare methods to calculate attributions. Second-order IG attributions can be computed using either the Hessian formulas \eqref{eq:mixedattr}, \eqref{eq:squaredattr}, or by the composition of attribution operators. First order attributions were calculated in three ways: one directly using the IG definition, and the other two by using the marginalization property over the above second order attributions. 
We use the right-hand rule to approximate the integral, using $M=100$ point quadrature. The Figure shows that these three approaches give approximately the same explanations, as expected by the marginalization property. We also observe that the interaction structure predicted by the second-order attributions agrees with the ground truth. We also consider the computation of third order attributions in Fig \ref{fig:synthetic}. In this case, we recover the interaction structure that we anticipated from \eqref{eq:synthpoly}.

\subsection{Real estate valuation}
To test our theory on a real dataset, we consider a real estate valuation dataset of housing prices in Taipei \cite{yeh2018building}. In this dataset, there are 416 observations of six numerical covariates (transaction date, house age, distance to the nearest metro station, number of nearby convenience stores, latitude, and longitude), and the target variable is the house price. To make predictions, we train a generalized linear model with a quadratic input function and a logit link function. 

We take three houses selected at random and compare explanations of the house valuation. We observe that, although the first order attributions are different because these houses have different characteristics, they are similar in the graphs they produce. Some features appear to act jointly, such as the distance to the metro, the number of stores, and latitude, whereas other features contribute in isolation. Without considering the specifics of house valuations in Taipei, the results suggest that higher order attribution analysis can reveal groups of features that act jointly during prediction.

\begin{figure}
    \centering
    \includegraphics[width=0.84\linewidth]{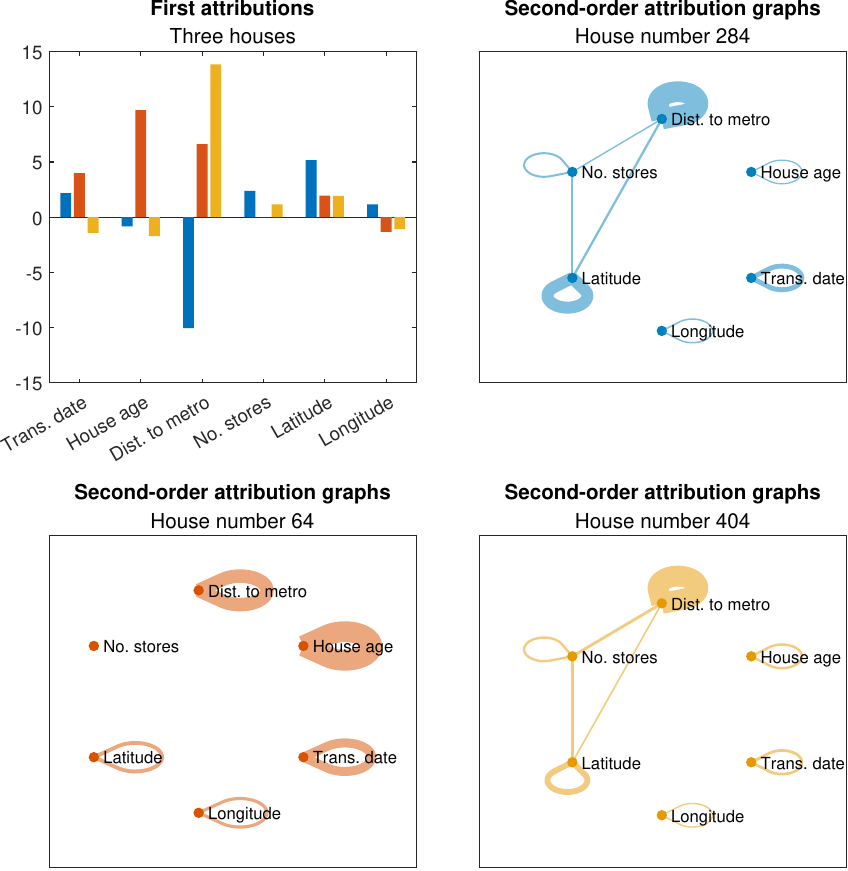}
    \caption{Explanations of house price valuations. Graphs are shown for the second-order attributions, where edge width corresponds to strength of attribution.}
    \label{fig:taipei}
\end{figure}

\section{Discussion and Conclusion}
\label{sec:discussion}
Recent works have analyzed the relationship between feature attribution and statistical interactions. Notably, the Integrated Hessians framework \cite{janizek2021explaining} coincides with our second-order case, but it does not consider higher order explanations. For the Shapley attribution framework, there are also  notions of Shapley interaction indices \cite{fumagalli2023shap}. We distinguish ourselves from these frameworks by defining attribution in terms of an operator theoretic framework, as opposed to scalar-valued scores on subsets of features. This preserves algebraic and compositional structure in a way that relates to combinatorial structures like graphs and simplicial complexes.

% \section{Conclusion}
\label{sec:conclusion}
In this work, we introduced a theory of higher-order attribution based on the composition of linear operators. This perspective exposes connections between interactions, feature attributions, and graphical representations that we believe are meaningful for future work. Our preliminary results  demonstrate the potential of this approach to uncover novel insights about complex models.

%-------------------------------------------------------------------------
\bibliographystyle{IEEEbib}
\bibliography{refs.bib} 
\label{sec:refs}

\end{document}